\documentclass[runningheads]{llncs}
\usepackage{graphicx}
\usepackage{multirow}
\usepackage{comment}
\usepackage{amsmath,amssymb} 
\usepackage{color}

\usepackage[pagebackref=true,breaklinks=true,linkcolor=red,anchorcolor=blue,            citecolor=green,letterpaper=true,colorlinks,bookmarks=false]{hyperref}

\begin{document}
\pagestyle{headings}
\mainmatter
\def\ECCVSubNumber{4860}  

\title{JGR-P2O: Joint Graph Reasoning based Pixel-to-Offset Prediction Network
for 3D Hand Pose Estimation from a Single Depth Image} 

\titlerunning{JGR-P2O for 3D Hand Pose Estimation from a Single Depth Image}
%
\author{Linpu Fang\inst{1} \and
Xingyan Liu\inst{1} \and
Li Liu\inst{2} \and
Hang Xu\inst{3} \and
Wenxiong Kang\inst{1}\thanks{Corresponding author: auwxkang@scut.edu.cn}
}
\authorrunning{Linpu Fang, Xingyan Liu et al.}
%
\institute{South China University of Technology, China \and
Center for Machine Vision and Signal Analysis, University of Oulu, Finland \and
Huawei Noah's Ark Lab
}
\maketitle

\begin{abstract}
State-of-the-art single depth image-based 3D hand pose estimation methods are based on dense predictions, including voxel-to-voxel predictions, point-to-point regression, and pixel-wise estimations. Despite the good performance, those methods have a few issues in nature, such as the poor trade-off between accuracy and efficiency, and plain feature representation learning with local convolutions. In this paper, a novel pixel-wise prediction-based method is proposed to address the above issues. The key ideas are two-fold: a) explicitly modeling the dependencies among joints and the relations between the pixels and the joints for better local feature representation learning; b) unifying
the dense pixel-wise offset predictions and direct joint regression for end-to-end training. Specifically, we first propose a graph convolutional network (GCN) based joint graph reasoning module to model the complex dependencies among joints and augment the representation capability of each pixel. Then we densely estimate all pixels' offsets to joints in both image plane and depth space and calculate the joints' positions
by a weighted average over all pixels' predictions, totally discarding the complex post-processing operations. The proposed model is implemented with an efficient 2D fully convolutional network (FCN) backbone and has only about 1.4M parameters. Extensive experiments on multiple 3D hand pose estimation benchmarks demonstrate that the proposed method
achieves new state-of-the-art accuracy while running very efficiently with around a speed of 110fps on a single NVIDIA 1080Ti GPU \footnote{This work was supported in part by the National Natural Science Foundation of China under Grants 61976095, in part by the Science and Technology Planning Project of Guangdong Province under Grant 2018B030323026. 
}. The code is available at \url{https://github.com/fanglinpu/JGR-P2O}.
\keywords{3D hand pose estimation, depth image, graph neural network}
\end{abstract}

\section{Introduction}
Vision-based 3D hand pose estimation aims to locate hand joints in 3D space from input hand images, which serves as one of the core techniques in contactless human computer interaction applications, such as virtual reality, augmented reality and robotic gripping \cite{mueller2017real,guleryuz2018fast}.
Recent years have witnessed significant advances \cite{du2019crossinfonet,moon2018v2v,yuan2018depth,madadi2017end,ge2016robust} in this area with the availability of consumer depth cameras, such as Microsoft Kinect and Intel RealSense, and the success of deep learning technology in the computer vision community. However, accurate and real-time 3D hand pose estimation is still a challenging task due to the high articulation complexity of the hand, severe self-occlusion
between different fingers, poor quality of depth images, etc.
\par In this paper, we focus on the problem of 3D hand pose estimation from a single depth image. At present, the state-of-the-art approaches to this task rely on deep learning technology, especially deep convolutional neural networks (CNNs). The main reasons are two-fold. On one hand,
public available large datasets \cite{yuan20172017,tang2014latent,tompson2014real,sun2015cascaded} with fully labeled 3D hand poses provide a large number of training data for these data-hungry methods. On the other hand, CNNs with well-designed network structures provide very effective solutions to challenging visual learning tasks and have been demonstrated to outperform traditional
methods by a large margin in various computer vision tasks, including 3D hand pose estimation.
\par Best performing deep learning-based methods are detection-based, which formulate 3D hand pose parameters as volumetric heat-maps or extended 3D heat-maps together with offset vector fields and estimate them in a dense prediction manner with fully convolutional networks (FCNs) or PointNet \cite{qi2017pointnet,qi2017pointnet++}. Contrary to their
regression-based counterparts that directly map the depth images to 3D hand pose parameters and severely suffer from the problem of highly non-linear mapping, the detection-based methods can learn better feature representations by pose reparameterization and have proven to be more effective for both human pose estimation \cite{pavlakos2017coarse,chu2017multi}
and hand pose estimation \cite{ge2018point,moon2018v2v,wan2018dense}.
\par By analyzing previous detection-based methods, we find that they suffer from several drawbacks in nature, which can be improved to boost performance. First, they bear the problem of poor trade-off between accuracy and efficiency. For example, the V2V \cite{moon2018v2v} uses 3D CNNs to estimate volumetric heat-maps, which is very parameter-heavy and computationally inefficient. The pixel-wise and point-wise prediction-based methods \cite{wan2018dense,ge2018point} take the advantages of 2D CNNs or PointNet to regress dense 3D estimations.
In spite of the higher efficiency, these methods achieve lower estimation precision empirically, and the complex post-processing operations still degrade the computational efficiency. Second, they consist of non-differentiable post-processing operations, such as taking maximum
and taking neighboring points, preventing fully end-to-end training and causing inevitable quantization errors. In addition, the models are trained with non-adaptive Gaussian heat-maps or joint-centered heat-maps, which may be suboptimal. Finally, the feature representation for each element (e.g., a voxel, a pixel or a point) is only learned by local convolutions ignoring the global context information. However, modeling the dependencies among joints and the relations between the elements and the joints helps to learn more abundant contextual information and better local feature representations.
\par To cope with these problems, we propose a novel joint graph reasoning based pixel-to-offset prediction network (JGR-P2O) for 3D hand pose estimation, which aims at directly regressing joints' positions from single depth images. Specifically, we decompose the 3D hand pose into joints' 2D image plane coordinates and depth values, and estimate these parameters in an ensemble way, fully exploiting the 2.5D property of depth images. The proposed method consists of two key modules, i.e., GCN-based joint graph reasoning module and pixel-to-offset prediction module. The joint graph reasoning module aims at learning a better feature representation for each pixel, which is vital for dense prediction. First, the features of joints are generated by summarizing the global information encoded in local features. Second, the dependencies among joints are modeled by graph reasoning to obtain stronger feature representations
of joints. Finally, the evolved joints' features are mapped back to local features accordingly enhancing the local feature representations. The pixel-to-offset prediction module densely estimates all the pixels' offsets to joints in both image plane and depth space. And the joints' positions in both image plane space and depth space are calculated by a weighted average over all the pixels' predictions. In this way, we discard the complex post-processing operations used in \cite{wan2018dense,ge2018point}, which improves not only the computational efficiency but also the estimation robustness.
\par Note that our JGR-P2O can obtain joints' positions directly from single depth images without extra post-processing operations. It generates intermediate dense offset vector fields, and can also be fully end-to-end trained under the direct supervision of joints' positions, fully sharing
the merits of both detection-based and regression-based methods. It also explicitly models the dependencies among joints and the relations between the pixels and the joints to augment the local feature representations. The whole model is implemented with an efficient 2D FCN backbone and has only about 1.4M parameters. It generally outperforms the previous detection-based methods on effectiveness and efficiency simultaneously. Overall, the proposed method provides some effective solutions to
the problems encountered by previous detection-based methods.
\par To sum up, the main contributions of this paper are as follows:
\begin{itemize}
\item We propose an end-to-end trainable pixel-to-offset module by leveraging the 2.5D property of depth images to unify the dense pixel-wise offset predictions and direct joint regression.
\item We propose a GCN-based joint graph reasoning module to explicitly model the dependencies among joints and the relations between the pixels and the joints to augment the local feature representations.
\item We conduct extensive experiments on multiple most common 3D hand pose estimation benchmarks (i.e., ICVL \cite{tang2014latent}, NYU \cite{tompson2014real},
and MSRA \cite{sun2015cascaded}). The results demonstrate that the proposed method achieves new state-of-the-art accuracy with only about 1.4M parameters while running very efficiently with around a speed of 110fps on single NVIDIA 1080Ti GPU.
\end{itemize}

\section{Related Work}
This paper focuses on the problem of estimating 3D hand pose from a single depth image. The approaches to this problem can be categorized into discriminative methods \cite{wan2016hand,sun2015cascaded,ge2018hand,ge20173d},
generative methods \cite{tzionas2016capturing,khamis2015learning,remelli2017low}
and hybrid methods \cite{tang2018opening,oberweger2015training,zhou2016model,ye2016spatial,wan2017crossing,oberweger2017deepprior++,oberweger2019generalized}.
In this section, we focus on the discussions of the deep learning-based discriminative and hybrid methods related closely to our work. These
methods can be further classified into regression-based methods, detection-based methods, hierarchical and structured methods. Please refer to \cite{yuan2018depth,supancic2015depth,supanvcivc2018depth}
for more detailed review. Furthermore, we also introduce some GCN-based
works that related to our method.
\par \textbf{Regression-based Methods. }Regression-based methods \cite{oberweger2015hands,oberweger2017deepprior++,ge2018hand,ge20173d,chen2018shpr,guo2017towards} aim at directly regressing 3D hand pose parameters such as 3D coordinates or joint angles. Oberweger et al. \cite{oberweger2017deepprior++,oberweger2015hands} exploit a bottleneck layer to learn a pose prior for constraining the hand pose. Guo et al. \cite{guo2017towards} propose a tree-structured Region Ensemble Network (REN) to regressing joints' 3D coordinates directly. Instead of using depth images as inputs, other works focus on 3D input representations, fully utilizing the depth information. Ge et al. \cite{ge20173d,ge2018hand} apply 3D CNNs and PointNet \cite{qi2017pointnet,qi2017pointnet++} for estimating 3D hand joint positions directly, which use 3D volumetric representation and 3D point cloud as inputs respectively. Despite the simplicity, the global regression manner within the fully-connected layers incurs highly non-linear mapping, which may reduce the estimation performance. However, our method adopts the dense prediction manner to regress the offsets from pixels to joints, effectively maintaining the local spatial context information.
\par\textbf{Detection-based Methods. }Detection-based methods \cite{ge2016robust,moon2018v2v,ge2018point,wan2018dense}
work in dense local prediction manner via setting a heat map for each
joint. Early works \cite{tompson2014real,ge2016robust} firstly detect
the joints' positions in 2D plain based on the estimated 2D heat-maps
and then translate them into 3D coordinates by complex optimization-based
post-processing. Recent works \cite{moon2018v2v,ge2018point,wan2018dense}
directly detect 3D joint positions from 3D heat-maps with much more
simple post-processing. Moon et al. \cite{moon2018v2v} propose a
Voxel-to-Voxel prediction network (V2V) for both 3D hand and human
pose estimation. Wan et al. \cite{wan2018dense} and Ge et al. \cite{ge2018point} formulate 3D hand pose as 3D heat-maps and unit vector fields and estimate these parameters by dense pixel-wise and point-wise regression respectively. Despite the good performance, these methods have some drawbacks, such as the poor trade-off between accuracy and efficiency and local feature representations. With the proposed pixel-to-offset prediction module and GCN-based joint graph reasoning module, our method can effectively solve these problems.

\begin{figure}
\centering
\includegraphics[scale=0.35]{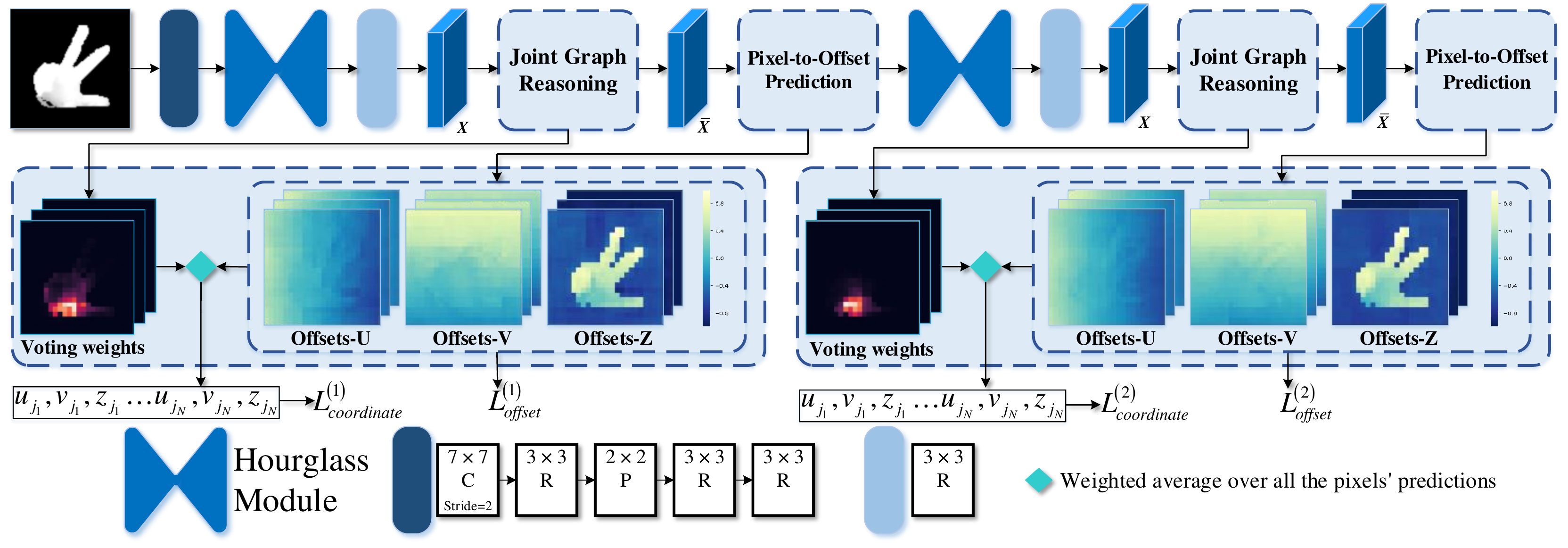}
\caption{\label{fig:Overview}An overview of our JGR-P2O. The abbreviations C, P, R indicate convolutional layer with BN and ReLU, pooling layer and residual module respectively. Given a hand depth image, the backbone module first extracts the intermediate local feature representation $\mathbf{\boldsymbol{\mathit{X}}}$, which is then augmented by the proposed GCN-based joint graph reasoning module producing the augmented local feature representation $\bar{\boldsymbol{X}}$. Finally, the proposed pixel-to-offset prediction module predicts three offset maps for each joint where each pixel value indicates the offset from the pixel to the joint along one of the axes in the UVZ coordinate system. The joint's UVZ coordinates are calculated as the weighted average over all the pixels' predictions. Two kinds of losses, coordinate-wise regression loss $L_{coordinate}$ and pixel-wise offset regression loss $L_{offset}$, are proposed to guide the learning process. We stack together two hourglasses to enhance the learning power, feeding the output from the previous module as the input into the next while exerting intermediate supervision at the end of each module.}
\end{figure}

\par \textbf{Hierarchical and Structured Methods. }These methods aim at
incorporating hand part correlations or pose constraints into the
model. Hierarchical methods \cite{madadi2017end,chen2019pose,du2019crossinfonet}
divide the hand joints into different subsets and use different network
branches to extract local pose features for each subset. Then all
the local pose features are combined together forming the global hand
pose representation for final pose estimation. Structured methods
\cite{oberweger2017deepprior++,oberweger2015hands,madadi2017end,zhou2016model} impose physical hand motion constraints into the model, which are
implemented by embedding constraint layers in the CNN model \cite{oberweger2017deepprior++,oberweger2015hands,zhou2016model}
or adding specific items in the loss function \cite{zhou2016model}.
Different from these methods, the proposed GCN-based joint graph reasoning module aims at augmenting the local feature representation by learning the dependencies among joints and the relations between the pixels and the joints.

\par \textbf{Related GCN-based Works. }Graph CNNs (GCNs) generalize CNNs
to graph-structured data. Approaches in this field are often classified
into two categories: spectral based methods \cite{defferrard2016convolutional,kipf2016semi}
that start with constructing the frequency filtering, and spatial
based methods \cite{monti2017geometric,velivckovic2017graph} that
generalize the convolution to a patch operator on groups of node neighbors. Recently, some works use GCNs for 3D pose estimation \cite{cai2019exploiting} and skeleton-based action regnition \cite{yan2018spatial,li2019actional,shi2019two}. And some works \cite{liang2018symbolic,li2018beyond} use GCN-based methods to augment the local feature representation for dense prediction. Inspired by these works, we also define the connections between hand joints as a graph and apply a GCN to learn their dependencies. Moreover, we design several different joint graph structures for comprehensive comparison studies.

\section{The Proposed Method}
\subsection{Overview}
\begin{figure}
\centering
\includegraphics[scale=0.4]{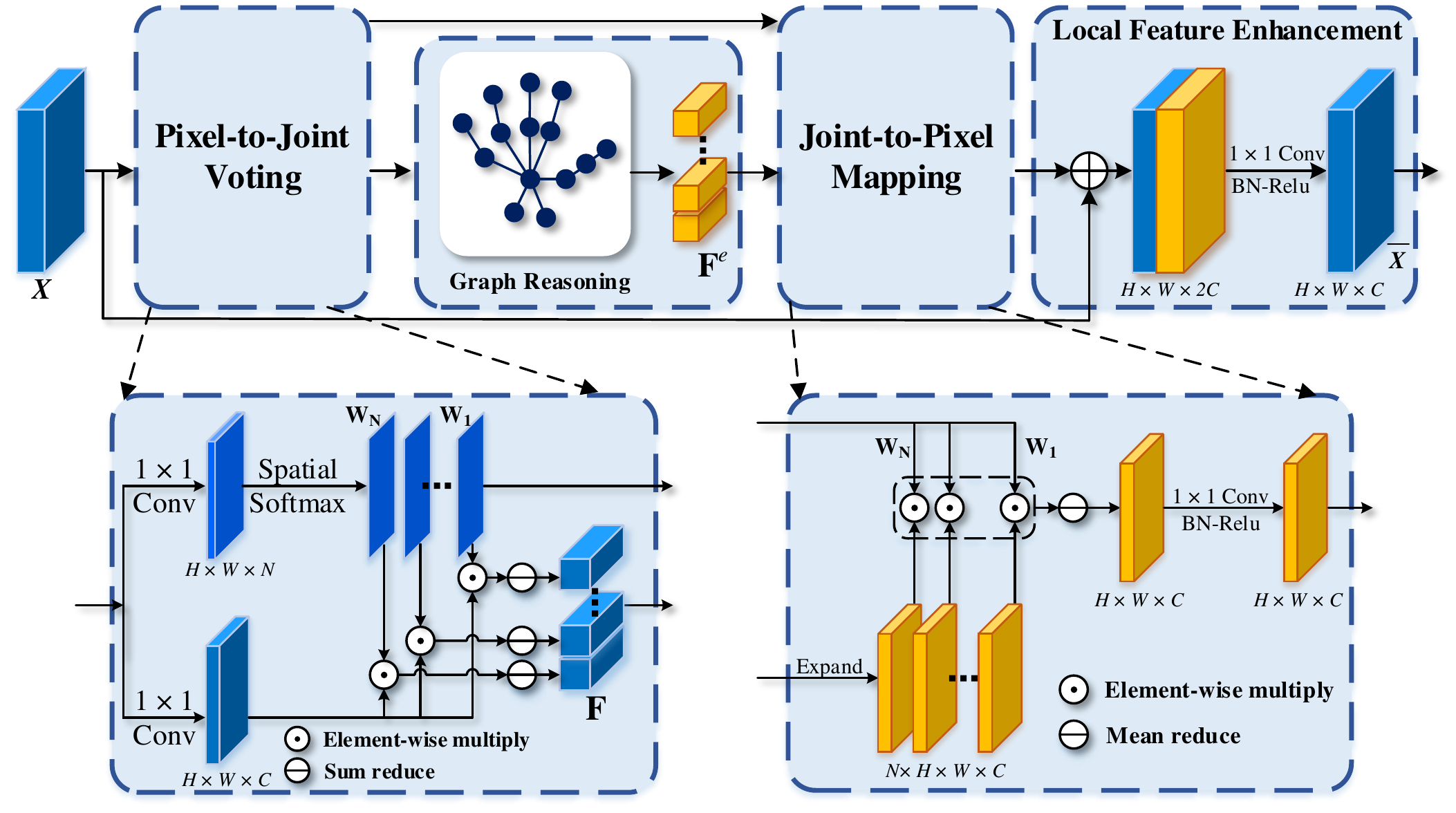}
\caption{\label{fig:JGR}Flowchart of our proposed GCN-based joint graph reasoning module. Given the intermediate feature representation $\boldsymbol{X}$ extracted from the backbone module, it first generates the joints' feature representation $\mathbf{F}$ by a pixel-to-joint voting mechanism where each joint is represented as the weighted average over all the local features. Then we define the connections between joints as a graph and map the joints' features to the corresponding graph nodes. The joints' features are propagated within the graph by graph reasoning, obtaining the enhanced joints' feature $\mathbf{F}^{\mathit{e}}$. Next, the $\mathbf{F}^{\mathit{e}}$ is mapped back to local features by a joint-to-pixel mapping mechanism that is the inverse operation of the pixel-to-joint voting, generating the joint context representations for all pixels. Finally, the original feature representation and the joint context representation are fused together obtaining enhanced local feature representation.}
\end{figure}

The proposed JGR-P2O casts the problem of 3D hand pose estimation
as dense pixel-to-offset predictions, fully exploiting the 2.5D property
of depth images. It takes a depth image as input and outputs the joints'
positions in image plain (i.e., uv coordinates) and depth space (i.e.,
z coordinates) directly. An overview of the JGR-P2O can be found in
Figure \ref{fig:Overview}. We use the high-efficient hourglass network
\cite{newell2016stacked} as the backbone to extract intermediate
local feature representation. Then the proposed joint graph reasoning
module models the dependencies among joints and the relations between
the pixels and the joints enhancing the intermediate local feature
representation. Finally, the pixel-to-offset module estimates the
offsets from pixels to joints and aggregates all the pixels' predictions
to obtain final joints' positions. Following \cite{newell2016stacked},
we stack together two hourglasses to enhance the learning power, feeding
the output from the previous module as the input into the next while
exerting intermediate supervision at the end of each module.

\subsection{GCN-based Joint Graph Reasoning Module}
Inspired by the symbolic graph reasoning (SGR) \cite{liang2018symbolic}, we propose the joint graph reasoning module to augment the intermediate local feature representation for each pixel, which is vital for local prediction. Given the extracted feature map from the backbone, we first generate the joints' features by summarizing the global context information encoded in local features. Specifically, joints are represented as the weighted average over all the local features through a pixel-to-joint voting mechanism. Then a joint-to-joint undirected graph $\mathbf{\mathit{\boldsymbol{G}}}=<\mathcal{N},\mathcal{\mathcal{E}}>$
is defined, where each node in $\mathcal{N}$ corresponds to a joint
and each edge $e_{i,j}\in\mathcal{\mathcal{E}}$ encodes relationship
between two joints. And the joints' features are propagated with the
defined structure of $\boldsymbol{G}$ to capture the dependencies
among joints and enhance their representation capabilities further.
Finally, the evolved joints' features are mapped back to local features
through a joint-to-pixel mapping mechanism, obtaining the pixel-wise
joint context representations which are combined with original local
features to enhance the local feature representations. The detailed
pipeline of this module can be found in Figure \ref{fig:JGR}.

\subsubsection{Pixel-to-Joint Voting}
We seek to obtain joints' visual representations based on the global
context information encoded in local features. Specifically, each
joint has its informative pixels, the representations of which are
aggregated to form the joint's feature. In this paper, we compute
the joints' features by a pixel-to-joint voting mechanism. Given the
feature map $\mathbf{\boldsymbol{\mathit{X}}}\in\mathbb{{R}}^{H\times W\times C}$ after the backbone network, where $H$, $W$ and $C$ denote the height, width and number of channels of the feature map respectively. First, the voting weights from pixels to joints are computed as:
\begin{equation}
\mathbf{\boldsymbol{\mathit{W}}}=\Phi\left(\phi\left(\boldsymbol{\mathbf{\mathit{X}}}\right)\right),
\end{equation}
where $\phi\left(\cdot\right)$ is a transformation function implemented
by a $1\times1$ convolution, $\Phi$ is the spatial softmax normalization,
and $\mathbf{\boldsymbol{\mathit{W}}}\in\mathbb{{R}}^{H\times W\times N}$
is the voting tensor where the $kth$ channel $\mathbf{W}_{k}\mathbf{\in\mathbb{{R}}^{\mathit{H\times W}}}$
represents the voting matrix for joint $k$. Then the feature representation for joint $k$ is calculated as the weighted average over all the transformed pixel-wise representations:
\begin{equation}
\mathbf{f_{\mathrm{\mathit{k}}}=\sum_{\mathrm{\mathit{i}}}\mathrm{\mathit{w}_{\mathit{ki}}\varphi\left(\mathbf{x}_{\mathit{i}}\right)}\mathrm{,}}
\end{equation}
where $\mathbf{x}_{\mathit{i}}$ is the representation of pixel $p_{i}$,
$\varphi\left(\cdot\right)$ is a transformation function implemented
by a $1\times1$ convolution layer, and $w_{\mathit{ki}}$, an element
of $\mathbf{W}_{k}$, is the voting weight for pixel $p_{i}$. We
also define the whole representation of all $N$ joints as $\mathbf{F}=\left[\mathbf{f}_{1}^{T};\ldots;\mathbf{f}_{N}^{T}\right]$.

\subsubsection{Graph Reasoning}
Given the joints' features and the defined joint-to-joint undirected
graph $\boldsymbol{G}$, it is natural to use a GCN to model the dependencies among joints and augment the joints' feature representations further. Following GCN defined in \cite{kipf2016semi}, we perform graph reasoning over representation $\mathbf{F}$ of all joints with matrix multiplication, resulting the evolved joint features $\mathbf{F^{\mathit{e}}}$:
\begin{equation}
\mathbf{F}^{e}=\sigma\left(\mathbf{A}^{e}\mathbf{F}\mathbf{W}^{e}\right),
\end{equation}
where $\mathbf{W}^{e}\in\mathbb{{R}}^{\mathit{C\times C}}$ is a trainable
transformation matrix, $\mathbf{A}^{e}\in\mathbb{{R}}^{N\times N}$
is the connection weight matrix defined according to the edge connections
in $\mathcal{\mathcal{E}}$, and $\sigma\left(\cdot\right)$ is a
nonlinear function (we use ReLU function for $\sigma\left(\cdot\right)$
in this paper). To demonstrate the generalization capability of the
GCN-based joint graph reasoning module, we try three different methods
to construct graph structure (i.e., the definition of $\mathbf{A}^{e}$)
in this paper.
\par \textbf{Skeleton Graph. }The most intuitive method is to define the
edge connections as hard weights (i.e., $\left\{ 0,1\right\} $) based
on the physical connections between joints in the hand skeleton. Then
the connection weight matrix is defined as the normalized form as
in \cite{kipf2016semi}: $\mathbf{A}^{e}=\tilde{\mathbf{D}}^{-\frac{1}{2}}\left(\mathbf{A}+\mathbf{I}_{N}\right)\tilde{\mathbf{D}}^{-\frac{1}{2}}$,
where $\mathbf{A}$ is the adjacency matrix defined in the hand skeleton,
$\mathbf{I}_{N}$ is the identity matrix, and $\mathbf{\tilde{A}=A}+\mathbf{I}_{N}$
defines a undirected graph with added self-connections. $\tilde{\mathbf{D}}$
is the diagonal node degree matrix of $\tilde{\mathbf{A}}$ with $\tilde{\mathbf{D}}_{ii}=\sum_{j}\tilde{\mathbf{A}}_{ij}$.
\par \textbf{Feature Similarity. }The connection weight between two joints can be calculated as the similarity of their visual representations:
$a_{ij}^{e}=\frac{exp\left(\upsilon\left(\mathbf{f}_{\mathrm{\mathit{i}}}\right)^{T}\psi\left(\mathbf{f}_{j}\right)\right)}{\ensuremath{\sum_{j=1}^{N}}exp\left(\upsilon\left(\mathbf{f}_{\mathrm{\mathit{i}}}\right)^{T}\psi\left(\mathbf{f}_{j}\right)\right)}$,
where $\upsilon$ and $\psi$ are two liner transformation functions
implemented by two fully-connected layers. Note that each sample has
a unique graph learned by this data-dependent method.
\par \textbf{Parameterized Matrix. }In this way, $\mathbf{A}^{e}$ is defined as a parameterized matrix whose elements are optimized together with the other parameters in the training process, that is, the graph is
completely learned according to the training data.

\subsubsection{Joint-to-Pixel mapping}
The evolved joint features can be used to augment the local feature
representations. Specifically, the pixel-wise joint context representations are first calculated by mapping the evolved joint features back to local features and then combined with original pixel-wise representations to compute the augmented local feature representations. We use the inverse operation of pixel-to-joint voting, i.e., joint-to-pixel mapping, to calculate the pixel-wise joint context representation. For pixel $p_{i}$, we first compute its context representation of joint $k$ as: $\mathbf{c}_{ik}=\mathit{w}_{\mathit{ik}}\mathbf{f}_{k}^{e}$,
where $\mathbf{f}_{k}^{e}$ is the evolved feature of joint $k$,
and $\mathit{w}_{\mathit{ik}}$ is the mapping weight from joint $k$
to pixel $p_{i}$, which is the same as the voting weight $\mathit{w}_{\mathit{ki}}$
in formula $\left(2\right)$. Then the mean of set $\left\{ \mathbf{c}_{ik};k=1,\ldots N\right\} $
is used to calculate the final pixel-wise joint context representation
for pixel $p_{i}$:
\begin{equation}
\mathbf{c}_{i}=\rho\left(\frac{1}{N}\sum_{k}\mathbf{c}_{ik}\right),
\end{equation}
where $\rho$ is a transformation function implemented by a $1\times1$
convolution with BN and ReLU.

\subsubsection{Local Feature Enhancement}
Finally, we aggregate the original feature representation $\mathbf{x}_{\mathit{i}}$
and joint context representation $\mathbf{c}_{i}$ to obtain the augmented
feature representation for pixel $p_{i}$:
\begin{equation}
\bar{\mathbf{x}}_{i}=\tau\left([\mathbf{c}_{i}^{T},\mathbf{x}_{\mathit{i}}^{T}]^{T}\right),
\end{equation}
where $\tau$ is a transformation function used to fuse the original
feature representation and joint context representation, and implemented
by a $1\times1$ convolution with BN and ReLU. The combination of
the augmented features of all the pixels constitutes the augmented
feature map $\bar{\boldsymbol{X}}$, which is used as the input to
the pixel-to-offset prediction module.

\subsection{Pixel-to-Offset Prediction Module}
A depth image consists of pixels' 2D image plane coordinates and depth
values (i.e., UVZ coordinates), which are the most direct information
for determining the positions of hand joints. In this paper, we also
decompose the 3D hand pose into joints' 2D image plane coordinates
and depth values, and estimate these parameters in an ensemble way.
More concretely, a pixel's UVZ coordinates and its offset vector to
a joint can determine the joint's position in the UVZ coordinate system.
That is, instead of predicting the joint's UVZ coordinates directly,
we can detour estimate the offset vector from the pixel to the joint
since the pixel's UVZ coordinates can be obtained from the depth image
directly. To achieve robust estimation, we aggregate the predictions
of all the pixels to obtain the position of the joint. Formally, for a certain joint $k$, we predict three offset values for each pixel representing the offset vector in the UVZ coordinate system from the pixel to joint $k$, resulting in three offset maps. Then the UVZ coordinates $\left(u_{j_{k}},v_{j_{k}},z_{j_{k}}\right)$
of joint $k$ is obtained by a weighted average over all the pixels'
predictions:
\begin{equation}
\begin{cases}
u_{j_{k}}=\sum_{i}w_{\mathit{ki}}\left(u_{p_{i}}+\Delta u_{ki}\right)\\
v_{j_{k}}=\sum_{i}w_{\mathit{ki}}\left(v_{p_{i}}+\Delta v_{ki}\right),\\
z_{j_{k}}=\sum_{i}w_{\mathit{ki}}\left(z_{p_{i}}+\Delta z_{ki}\right)
\end{cases}
\end{equation}
where $\left(u_{p_{i}},v_{p_{i}},z_{p_{i}}\right)$ indicate the UVZ
coordinates of pixel $p_{i}$, $\left(\Delta u_{ki},\Delta v_{ki},\Delta z_{ki}\right)$
represent the predicted offset values from pixel $p_{i}$ to joint
$k$. $w_{\mathit{ki}}$ is the normalized prediction weight of pixel
$p_{i}$, indicating its importance for locating the joint $k$, which
is set to be same as the voting weight introduced in Section 3.2.1.
The pixel-to-offset prediction module is implemented by a $1\times1$
convolution layer that takes the augmented local feature representation
$\bar{\boldsymbol{X}}$ as input and output $3N$ offset maps for
all the $N$ joints directly.
\par Note that our P2O module is much simpler than the estimation scheme of A2J \cite{xiong2019a2j} where two different branches are design to estimate the joints' UV coordinates and Z coordinates, respectively. In addition, A2J uses a single feature in high-level feature maps to predict multiple estimations for a set of anchor points, which may distract the model's representation learning as well as increasing the parameters. The experimental results also demonstrates the superiority of our method over the A2J.  

\begin{figure}
\centering
\includegraphics[scale=0.57]{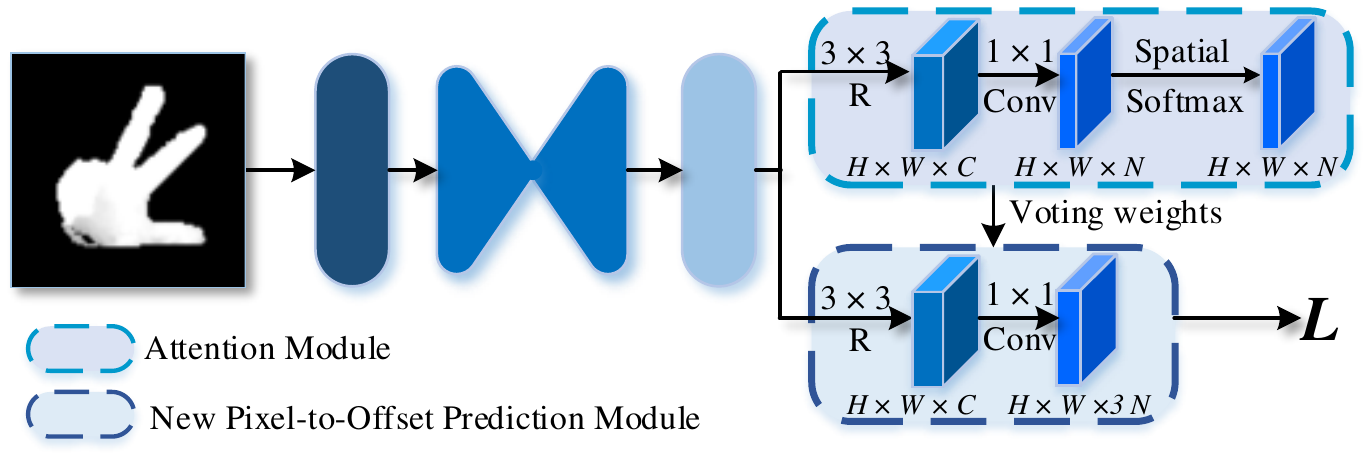}
\centering{}\caption{\label{fig:baseline_wo_graph}The flowchart of the attention-based
baseline model. The JGR module is replaced with an attention module
to calculate the weights of pixels for locating the joints. The attention
module consists of a $3\mathrm{x3}$ Residual block, a $1\mathrm{x1}$
Conv layer, and a spatial softmax operation. We also add a $3\mathrm{x3}$
Residual block to the original pixel-to-offset prediction module.
This figure only depicts the one-stage version of the attention-based
baseline model. In practice, we employ the two-stage version for comparison.}
\end{figure}

\subsection{Training Strategy}
The predicted joints' coordinates are calculated as in formula $\left(6\right)$. According to the ground truth 3D hand pose, we can construct a coordinate-wise regression loss:
\begin{equation}
L_{coordinate}=\sum_{k}\sum_{c}L_{\delta}\left(c_{j_{k}}-c_{j_{k}}^{*}\right),
\end{equation}
where $c_{j_{k}}$ is one of the predicted UVZ coordinates of joint
$k$, and $c_{j_{k}}^{*}$ is the corresponding ground truth coordinate.
We choose the Huber loss function $L_{\delta}$ as the regression
loss function since it is less sensitive to outliers in data than
squared error loss function. Moreover, we can also explicitly supervise
the generation process of offset maps by constructing a pixel-wise
offset regression loss:
\begin{equation}
L_{offset}=\sum_{k}\sum_{i}\sum_{c}L_{\delta}\left(\Delta c_{ki}-\Delta c_{ki}^{*}\right),
\end{equation}

where $\Delta c_{ki}$ is the offset value from pixel $p_{i}$ to
joint $k$ along one of axes in the UVZ coordinate system, and $\Delta c_{ki}^{*}$ is the corresponding ground truth offset value. The pixel-wise offset regression loss can be seen as a regularization term for learning better local feature representation. Note that we normalize the ground truth coordinates and offset values to be within the range $[-1,1]$, the pixel's UV coordinates and Z coordinates (i.e., depth values) are also normalized to be within the range $[0,1]$ and $[-1,1]$ respectively. Therefore, the estimated offset maps and joints' coordinates are also the normalized versions. We use a downsampled input depth image with the same resolution as the predicted offset map to calculate these parameters.
Following \cite{newell2016stacked}, we boost the learning capability
of the network architecture by stacking multiple hourglasses with
identical structures, feeding the output from the previous module
as the input into the next while exerting intermediate supervision
at the end of each module. The final loss for the whole network is
defined as follows:
\begin{equation}
L=\ensuremath{\sum_{s=1}^{S}} L_{coordinate}^{\left(s\right)}+\beta L_{offset}^{\left(s\right)}.
\end{equation}
where $L_{coordinate}^{\left(s\right)}$ and $L_{offset}^{\left(s\right)}$
are the coordinate-wise regression loss and pixel-wise offset regression
loss at the $sth$ stage, $\text{\ensuremath{\beta=0.0001}}$
is the weight factor for balancing the proposed two kinds
of losses, and $S=2$ is the total number of the stacked hourglasses.
The whole network architecture is trained in an end-to-end style with
the supervision of this loss.

\begin{table}
\caption{\label{tab:graph_structure}Comparison of different graph structures in the proposed JGR module. \#Params indicates the number of parameters of the whole model.}
\centering{}\tabcolsep 0.1in%
\begin{tabular}{ccc}
\hline 
{Graph Structures} & Mean error (mm) & {\#Params}\\
\hline 
Skeleton Graph & \textbf{8.29} & 1.37M\\
Feature Similarity & 8.45 & 1.43M\\
Parameterized Matrix & 8.36 & 1.37M\\
\hline 
\end{tabular}
\end{table}

\begin{table}
\centering{}\caption{\label{tab:ablation_studies}Effectiveness of individual components of the proposed method.}
\tabcolsep 0.1in%
\begin{tabular}{ccccc}
\hline 
\multicolumn{3}{c}{Component} & \multirow{2}{*}{Mean error (mm)}\\
\cline{1-3} 
P2O & Offset Loss & JGR & \\
\hline 
\checkmark &  &  & 10.83\\
\checkmark & \checkmark &  & 10.54$^{-0.29}$\\
\checkmark & \checkmark & \checkmark & \textbf{8.29$^{-2.25}$}\\
\hline 
\end{tabular}
\end{table}

\section{Experiments}
\subsection{Datasets and Settings}
We evaluate our proposed JGR-P2O on three common 3D hand pose estimation datasets: ICVL dataset \cite{tang2014latent}, NYU dataset \cite{tompson2014real}, and MSRA dataset \cite{sun2015cascaded}. The ICVL dataset contains 330K training and 1.5K testing depth images that are captured with an Intel Realsense camera. The ground truth hand pose of each image consists of $N=16$ joints. The NYU dataset was captured with three Microsoft Kinects from different views. Each view consists of 72K training and 8K testing depth images. There are 36 joints in each annotated hand pose. Following most previous works, we only use view 1 and $N=14$ joints for training and testing in all experiments. The MSRA dataset consists of 76K training images captured from 9 subjects with 17 gestures, using Intel\textquoteright s Creative Interactive Camera. Each image is annotated with a hand pose with $N=21$ joints. We use the leave-one-subject-out cross-validation strategy \cite{sun2015cascaded} for evaluation.
\par We employ two most commonly used metrics to evaluate the performance of 3D hand pose estimation. The first one is the mean 3D distance error (in mm) averaged over all joints and all test images. The second one is the percentage of success frames in which the worst joint 3D distance error is below a threshold.
\par All experiments are conducted on a single server with four NVIDIA 1080Ti GPU using Tensorflow. For inputs to the JGR-P2O, we crop a hand area from the original image using a method similar to the one proposed in \cite{oberweger2017deepprior++} and resize it to a fixed size of 96x96. The depth values are normalized to [-1, 1] for the cropped image. For training, Adam with weight decay of 0.00005 and batch size of 32 is used to optimize all models. Online data augmentation is used, including in-plane rotation ([-180, 180] degree), 3D scaling ([0.9, 1,1]), and 3D translation ([-10, 10] mm). The initial learning rate is set to be 0.0001, reduced by a factor of 0.96 every epoch. We train 8 epochs for the ICVL training set and 58 epochs for the other training sets.

\begin{table}
\caption{\label{tab:Hourglasses}Comparison of different numbers of stacked hourglass module.}
\centering{}\tabcolsep 0.1in%
\begin{tabular}{ccc}
\hline 
{\#Hourglasses} & Mean error (mm) & {\#Params}\\
\hline 
1 & 8.63 & 0.72M\\
2 & 8.29 & 1.37M\\
3 & 8.27 & 2.02M\\
\hline 
\end{tabular}
\end{table}

\begin{table}
\caption{\label{tab:baselines}Comparison with different baselines on NYU.}
\centering{}\tabcolsep 0.1in%
\begin{tabular}{ccc}
\hline 
{Model} & Mean error (mm) & {\#Params}\\
\hline 
Baseline with Attention Module & 8.72 & 1.42M\\
Baseline with DHM Module & 8.69 & 1.37M\\
Ours & \textbf{8.29} & 1.37M\\
\hline
\end{tabular}
\end{table}

\subsection{Ablation Studies }
We firstly conduct ablation studies to demonstrate the effectiveness of various components of the proposed JGR-P2O. The ablation studies are conducted on the NYU dataset since it is more challenge than the other two.

\textbf{Comparison of different graph structures.} Table \ref{tab:graph_structure} reports the performance of different graph structures in the proposed joint graph reasoning module. It can be seen that different graph structures can obtain similar estimation precision, indicating that the proposed joint graph reasoning module has strong generalization capability. In the following experiments, we choose the skeleton graph as the default graph structure for the joint graph reasoning module since it is more interpretable and best-performed.
\par \textbf{Effectiveness of individual components.} The results in Table \ref{tab:ablation_studies} show how much each component improves the estimation performance along with the combinations of other components. The simplest baseline that combines the backbone network and a P2O module, denoted as P2O in Table \ref{tab:ablation_studies}, estimates the joint's positions with the average summation over all pixels' predictions and obtains highest estimation error. Adding the pixel-wise offset regression loss for training decreases estimation error by 0.29mm. Finally, the JGR module helps to greatly decrease the estimation error by 2.25mm.    
\par \textbf{Number of hourglass modules. }The results of using different numbers of hourglass modules are reported in Table \ref{tab:Hourglasses}. It can be seen that with only one hourglass, the proposed JGR-P2O would
achieve relatively low mean 3D distance errors (8.63mm)
on the NYU dataset. Increasing the number of hourglasses
can improve the estimation precision, but three hourglasses can only obtain negligible improvement. In this paper, we stack only two hourglasses to balance accuracy and efficiency.

\begin{figure}
\centering
\includegraphics[scale=0.2]{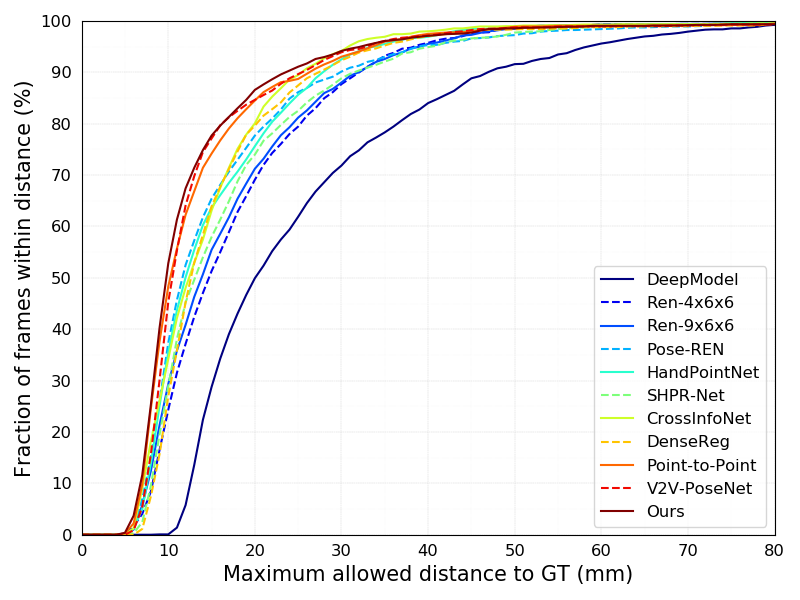}\includegraphics[scale=0.2]{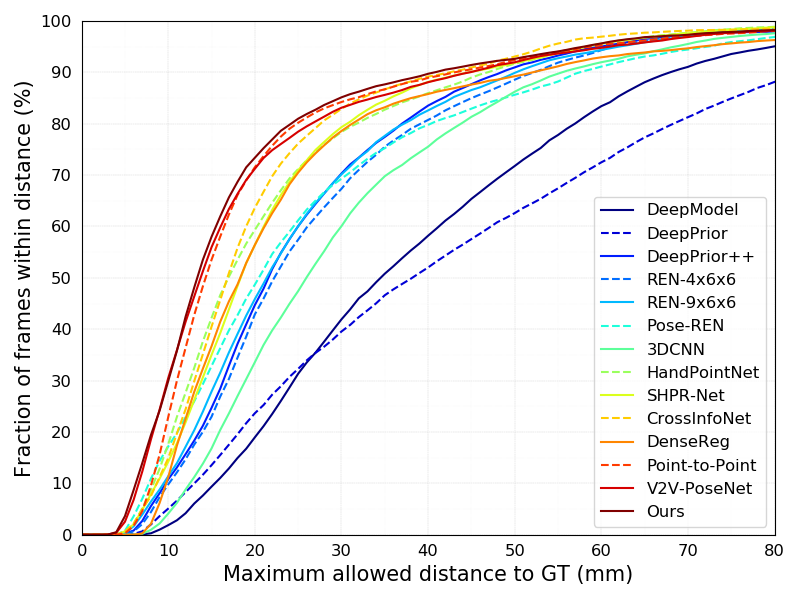}\includegraphics[scale=0.2]{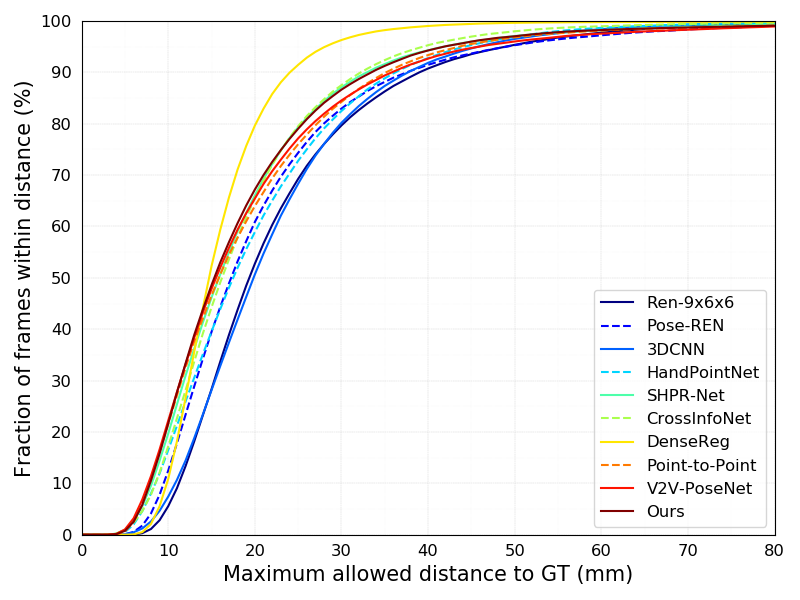}
\caption{\label{fig:comparisons}Comparison with previous state-of-the-art
methods. The percentages of success frames over different error thresholds
are presented in this figure. Left: ICVL dataset, Middle: NYU dataset,
Right: MSRA dataset.}
\end{figure}

\subsection{Comparison with different baselines}
To demonstrate the effectiveness of the proposed JGR and P2O module, we compare them with related baseline methods. The results are shown in Table \ref{tab:baselines}.
\par \textbf{JGR vs. Attention. }To verify the effectiveness of the proposed JGR module, we design an attention-based baseline model where the JGR module is replaced with an attention module to calculate
the weights of pixels for locating the joints. The flowchart of the baseline model can be found in Figure \ref{fig:baseline_wo_graph}. As shown in Table \ref{tab:baselines}, the JGR module outperforms the attention module by reducing the mean 3D distance error by 0.43mm on the NYU dataset, while having fewer parameters. It demonstrates that the JGR module is indeed useful for better local feature learning.
\par \textbf{P2O vs. Differentiable heat-map (DHM). }To demonstrate the effectiveness of the proposed P2O module, we compare our model with a model by replacing the P2O module with the DHM module proposed in \cite{iqbal2018hand}. DHM implicitly learns the joints' depth maps and heatmap distributions, while our P2O explicitly estimates the offsets from pixels to joints. It can be seen from Table \ref{tab:baselines} that our P2O module surpasses the DHM module by reducing the mean 3D distance error from 8.69mm to 8.29mm on NYU, which demonstrates the superiority of the proposed P2O module.

\subsection{Comparison with state-of-the-art}
We compare our proposed JGR-P2O with state-of-the-art deep learning-based
methods, including both dense prediction-based methods: dense regression
network (DenseReg) \cite{wan2018dense}, Point-to-Point \cite{ge2018point},
Point-to-Pose Voting \cite{li2019point}, A2J \cite{xiong2019a2j},
and V2V \cite{moon2018v2v}, and direct regression-based methods:
model-based method (DeepModel) \cite{zhou2016model}, DeepPrior \cite{oberweger2015hands},
improved DeepPrior (DeepPrior++) \cite{oberweger2017deepprior++},
region ensemble network (Ren-4x6x6 and Ren-9x6x6 \cite{guo2017towards}),
Pose-guided REN (Pose-Ren) \cite{chen2019pose}, 3DCNN \cite{ge20173d},
HandPointNet \cite{ge2018hand} , SHPR-Net \cite{chen2018shpr} and
CrossInfoNet \cite{du2019crossinfonet}. The percentages of success
frames over different error thresholds and mean 3D distance errors
are shown in Figure \ref{fig:comparisons} and Table \ref{tab:Comparisons},
respectively.

\begin{table}
\centering{}\caption{\label{tab:Comparisons}Comparison with previous state-of-the-art methods on the ICVL, NYU and MSRA dataset. Mean error indicates the average 3D distance error. Type DR and DP indicate the direct regression-based method and dense prediction-based method, respectively. \#Params indicates the parameter quantity of the whole network. Speed indicates the running speed during testing.}
\tabcolsep 0.08in \renewcommand{\arraystretch}{1.0}%
\begin{tabular}{ccccccc}
\hline 
\multirow{2}{*}[0.001cm]{{\small{}Method}} & \multicolumn{3}{c}{{\small{}Mean error (mm)}} & \multirow{2}{*}{{\small{}Type}} & \multirow{2}{*}{{\small{}\#Params}} & {\small{}Speed}\tabularnewline
\cline{2-4} \cline{3-4} \cline{4-4} 
 & \multicolumn{1}{c}{{\small{}ICVL}} & {\small{}NYU} & {\small{}MSRA} &  &  & {\small{}(fps)}\tabularnewline
\hline 
{\small{}DeepModel\cite{zhou2016model}} & {\small{}11.56} & {\small{}17.04} & {\small{}-} & {\small{}DR} & {\small{}-} & {\small{}-}\tabularnewline
{\small{}DeepPrior\cite{oberweger2015hands}} & {\small{}10.40} & {\small{}19.73} & {\small{}-} & {\small{}DR} & {\small{}-} & {\small{}-}\tabularnewline
{\small{}DeepPrior++\cite{oberweger2017deepprior++}} & {\small{}8.10} & {\small{}12.24} & {\small{}9.50} & {\small{}DR} & {\small{}-} & {\small{}30.0}\tabularnewline
{\small{}REN-4x6x6\cite{guo2017towards}} & {\small{}7.63} & {\small{}13.39} & {\small{}-} & {\small{}DR} & {\small{}-} & {\small{}-}\tabularnewline
{\small{}REN-9x6x6\cite{guo2017towards}} & {\small{}7.31} & {\small{}12.69} & {\small{}9.70} & {\small{}DR} & {\small{}-} & {\small{}-}\tabularnewline
{\small{}Pose-REN\cite{chen2019pose}} & {\small{}6.79} & {\small{}11.81} & {\small{}8.65} & {\small{}DR} & {\small{}-} & {\small{}-}\tabularnewline
{\small{}3DCNN\cite{ge20173d}} & {\small{}-} & {\small{}14.1} & {\small{}9.60} & {\small{}DR} & {\small{}104.9M} & {\small{}215}\tabularnewline
{\small{}HandPointNet\cite{ge2018hand}} & {\small{}6.94} & {\small{}10.54} & {\small{}8.50} & {\small{}DR} & {\small{}2.58M} & {\small{}48.0}\tabularnewline
{\small{}SHPR-Net\cite{chen2018shpr}} & {\small{}7.22} & {\small{}10.78} & {\small{}7.76} & {\small{}DR} & {\small{}-} & {\small{}-}\tabularnewline
{\small{}CrossInfoNet\cite{du2019crossinfonet}} & {\small{}6.73} & {\small{}10.08} & {\small{}7.86} & {\small{}DR} & {\small{}23.8M} & {\small{}124.5}\tabularnewline
\hline 
{\small{}DenseReg\cite{wan2018dense}} & {\small{}7.30} & {\small{}10.2} & \textbf{\small{}7.20} & {\small{}DP} & {\small{}5.8M} & {\small{}27.8}\tabularnewline
{\small{}Point-to-Point\cite{ge2018point}} & {\small{}6.30} & {\small{}9.10} & {\small{}7.70} & {\small{}DP} & {\small{}4.3M} & {\small{}41.8}\tabularnewline
{\small{}V2V-PoseNet\cite{moon2018v2v}} & {\small{}6.28} & {\small{}8.42} & {\small{}7.59} & {\small{}DP} & {\small{}457.5M} & {\small{}3.5}\tabularnewline
{\small{}Point-to-Pose Voting\cite{li2019point}} & {\small{}-} & {\small{}8.99} & {\small{}-} & {\small{}DP} & {\small{}-} & {\small{}80.0}\tabularnewline
{\small{}A2J\cite{xiong2019a2j}} & {\small{}6.46} & {\small{}8.61} & {\small{}-} & {\small{}DP} & {\small{}44.7M} & {\small{}105.1}\tabularnewline
{\small{}JGR-P2O(Ours)} & \textbf{\small{}6.02} & \textbf{\small{}8.29} & {\small{}7.55} & {\small{}DP} & {\small{}1.4M} & {\small{}111.2}\tabularnewline
\hline 
\end{tabular}
\end{table}

It can be seen that dense prediction-based methods are generally superior
to direct regression-based methods. As shown in Table \ref{tab:Comparisons}, our method can achieve the lowest mean estimation errors (6.02mm and 8.29mm) on the ICVL and NYU dataset. Figure \ref{fig:comparisons} also shows that the proportions of success frames of our method are highest when the error thresholds are lower than 30mm and 50mm on the ICVL and NYU dataset, respectively. Our method obtains the second-lowest estimation error (7.55mm) on the MSRA dataset, which is only 0.35mm higher than the estimation error (7.20mm) of DenseReg \cite{wan2018dense}. 
\par Table \ref{tab:Comparisons} also shows that our method has the minimum
model size and fastest running speed, compared with state-of-the-art
dense prediction-based methods. Specifically, the total parameter
quantity of our network is only 1.4M, and the running speed of our
method is 111.2fps, including 2.0ms for reading and pre-processing
image, and 7.0ms for network inference on a NVIDIA 1080Ti GPU.
\par More experimental analysis including qualitative results can be found in the supplementary material. 

\section{Conclusions}
In this work, we propose a new prediction network (JGR-P2O) for 3D
hand pose estimation from single depth images. Within JGR-P2O the
GCN-based joint graph reasoning module can help to learn better local
feature representation by explicitly modeling the dependencies among
joints and the relations between pixels and joints, and the pixel-to-offset
prediction module unifies the dense pixel-wise offset predictions
and direct joint regression for end-to-end training, fully exploiting
the 2.5D property of depth images. Extensive experiments demonstrate
the superiority of the JGR-P2O concerning for both accuracy and efficiency.

%
%
\bibliographystyle{splncs04}
\bibliography{HPE}
\end{document}